\documentclass[10pt,journal,compsoc]{IEEEtran}
\usepackage{float, stfloats}
\usepackage{graphicx, color}
\usepackage{subfigure}
\usepackage{ragged2e}
\usepackage{amsmath}
\usepackage{indentfirst}
\usepackage{multirow, array}
\usepackage{colortbl}
\usepackage{booktabs}

\ifCLASSOPTIONcompsoc
  \usepackage[nocompress]{cite}
\else
  \usepackage{cite}
\fi

\begin{document}
%
\title{Learning Quintuplet Loss for Large-scale Visual Geo-Localization}

\author{Qiang Zhai\textsuperscript{1},
        Rui Huang\textsuperscript{1},
        Hong Cheng\textsuperscript{1},~\IEEEmembership{Senior Member,~IEEE,}
        Huiqin Zhan\textsuperscript{1},
        Jun Li\textsuperscript{2}
        and Zicheng Liu\textsuperscript{3},~\IEEEmembership{Fellow,~IEEE}
\IEEEcompsocitemizethanks{
\IEEEcompsocthanksitem Q. Zhai, R. Huang, H. Cheng, H. Zhan are with the Center for Robotics, UESTC, China, qiang.zh6@gmail.com; hcheng@uestc.edu.cn; ruihuang@uestc.edu.cn; zhanhq@uestc.edu.cn.
\IEEEcompsocthanksitem J. Li is with school of vehicle and transportation, Tsinghua University, China, lijun1958@tsinghua.edu.cn.
\IEEEcompsocthanksitem Z. Liu is with Microsoft Research, Redmond, USA, zliu@microsoft.com}
}

\markboth{}%
{Shell \MakeLowercase{\textit{et al.}}: Bare Demo of IEEEtran.cls for Computer Society Journals}

\IEEEtitleabstractindextext{%
\begin{abstract}
\justifying
With the maturity of Artificial Intelligence (AI) technology, Large Scale Visual Geo-Localization (LSVGL) is increasingly important in urban computing, where the task is to accurately and efficiently recognize the geo-location of a given query image. The main challenge of LSVGL faced by many experiments due to the appearance of real-word places may differ in various ways. While perspective deviation almost inevitably exists between training images and query images because of the arbitrary perspective. To cope with this situation, in this paper, we in-depth analyze the limitation of triplet loss which is the most commonly used metric learning loss in state-of-the-art LSVGL framework and propose a new QUInTuplet Loss (QUITLoss) by embedding all the potential positive samples to the primitive triplet loss. Extensive experiments have been conducted to verify the effectiveness of the proposed approach and the results demonstrate that our new loss can enhance various LSVGL methods.
\end{abstract}

}

\maketitle

\IEEEdisplaynontitleabstractindextext

\IEEEpeerreviewmaketitle

\IEEEraisesectionheading{
\section{Introduction}\label{sec:introduction}}
\IEEEPARstart{L}{arge} Scale Visual Geo-Localization (LSVGL) has been thought of as a key factor in automotive driving, augmented reality, robotics navigation, etc, and the past decade has seen the rapid development of LSVGL in computer vision [1, 2] and robotics communities [3, 4]. LSVGL has become a very important technical means of urban computing [5, 6]. Despite its long success, there is an increasing concern in LSVGL due to the varieties in the real world.

LSVGL has always been treated as an instance retrieval task, which proposes to find the most similar image by querying a large trained geo-tagged database, and almost all the LSVGL systems in recent years try to investigate the most satisfying network architecture to extract a more discriminate image representation. Two steps are generally needed to achieve the geometric location: 1). Training a large scale geo-tagged database; 2). Extracting the descriptor and estimating the optimal matching. 
\begin{figure}
	\centering
	\includegraphics[scale=0.35]{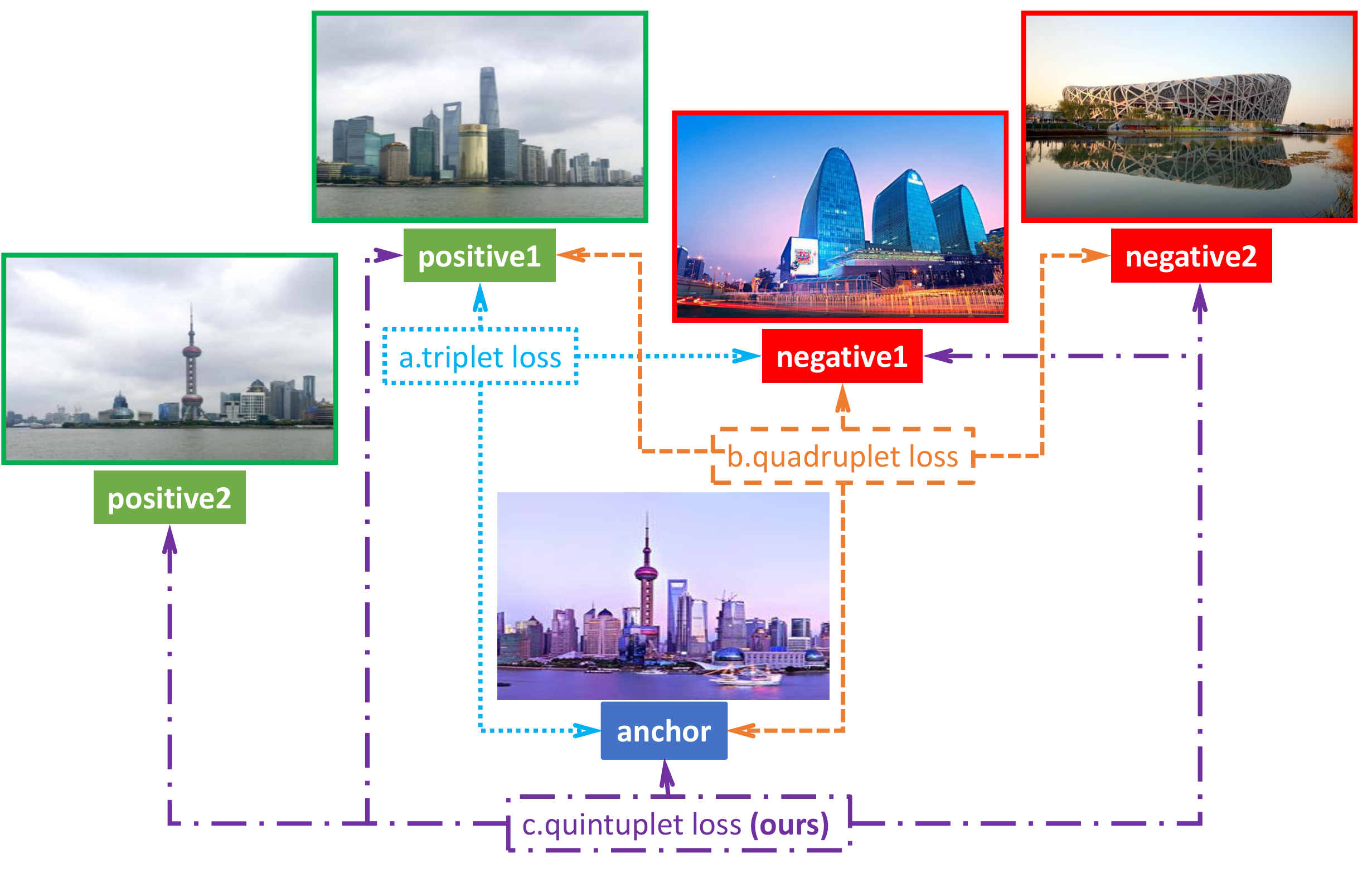}
	\caption{a. The classic triplet loss selects a positive sample and a negative sample. During the optimization process, the optimizer makes the anchor as close as possible to the positive sample and away from the negative sample. b. On the basis of triplet loss, the quadruplet loss selects an additional negative sample to make the anchor farther away from all the negative samples. c. We propose a new loss called quintuplet loss specifically for the LSVGL task. Different from the classic triplet loss and quadruplet loss, two positive samples are selected for the quintuplet loss (two are selected in the diagram, we will prove that it is best to choose two positive samples, both theoretically and experimentally) to strengthen the positive-anchor constraints while retaining the quadruplet loss selection strategy to pick negative samples.}
	\label{fig:cmp_with_other_loss}
\end{figure}

The sophisticated hand-crafted local invariant feature extractors contributed significantly to LSVGL in the early stages, such as  Scale-Invariant Feature Transformation (SIFT) and Speed-up Robust Features (SURF), and some global image features, as GIST, bag-of-words approaches. These traditional feature extraction techniques combined with machine learning algorithms such as SVM and AdaBoost have achieved remarkable results in many fields.

In recent years, with the enhancement of GPU-based computation power and data volume, Convolutional Neural Networks (CNNs) have greatly promoted the development of visual recognition, and have achieved state-of-the-art in many areas of recognition tasks such as object detection, scene classification, and image recognition. The principle of CNN has been developed for nearly 40 years since it was proposed in the 1980s. The most urgent research shows that the general features extracted by CNN have strong generalization ability. The features extracted in a certain field can be transferred and well generalized to other visual recognition tasks. Shallow features such as SIFT can easily cause the semantic gap problem in LSVGL, that is, features extracted from different locations may be very similar. Deep feature vectors extracted using CNN networks with deep layers stacking can effectively alleviate this problem.

Triplet loss was initially proposed in \cite{facenetSchroffKP15} to learn robust facial embedding and online triple mining. Different from the traditional softmax loss, triplet loss combines the anchor sample, positive sample, negative sample to form a tuple and helps the network to learn a kind of face embedding, so that minimizing the distance between the anchor face and positive samples, and maximizing the distance between the anchor and negative samples in the embedding space. Owing to the success of triplet loss, the loss derived from triplet loss has also been widely concerned \cite{quadruplet17, trihard17}. TriHard loss \cite{trihard17} selects the negative sample closest to the training sample as hard sample and uses hard sample to replace randomly selected negative sample in triplet loss. Quadruplet loss \cite{quadruplet17} is trained with two negative sample constraints, and the calculation of relative distance in triplet loss is modified to absolute distance. Margin Sample Mining Loss (MSML) \cite{mlms17} takes advantage of TriHard loss and quadruplet loss. After great success in the field of face recognition, researchers began to try to apply triplet loss to LSVGL and also achieved enormous great success \cite{arandjelovic2016netvlad, zhu2018attention, liuliu19}.

Unlike most of the work dedicated to proposing network architectures, we try to use existing network models to analyze the objective situation of the problem and explore whether the problem can be optimized from the perspective of loss. We believe that the biggest drawback of triplet loss and quadruplet loss is that they focus too much on maximizing the distance between anchor samples and negative samples, while ignoring the strong constraints between anchor samples and positive samples. Generally, due to the time difference between the anchor samples and the positive samples and the arbitrariness of the spatial position, although the image at the same location inevitably has a perspective deviation, that is, we can reach a consensus: an anchor sample consists of multiple positive samples in the location partial image composition, which is a significant difference between LSVGL and face recognition. Regardless of triple loss or quadruple loss, only one positive sample is considered, and we believe that this loss is not completely suitable for the LSVGL scenario. We put forward the hypothesis that adding all potential positive samples to the loss to train together will have a gain effect on LSVGL.

In this paper, we propose a new loss called QUInTuplet loss (QUITLoss), which attempts to maintain the advantages of the triple loss and adds all potential positive samples to the loss. This is because there is usually a difference in perspective between the anchor sample and the positive sample, but this problem does not exist in the field of facial recognition. We start with this general common sense and theoretically analyze the rationality of QUITLoss. After a careful comparison, we concluded that our QUITLoss could make up for the shortcomings of previous methods. Fig. \ref{fig:cmp_with_other_loss} shows the main idea of QUITLoss.

We verified the role of QUITLoss in experiments. We corrected the loss in the previous method with QUITLoss and performed ablation experiments. The experimental results verify the effectiveness of our method for the LSVGL task.

In the following, we overview the main contents of our method and summarize the contributions:

1). We embed several metric learning losses into state-of-the-art LSVGL framework and make up for the gap in the current LSVGL solutions that only use triplet loss and do not compare with other metric learning methods.

2). We propose a new metric learning loss strategy to compensate for the precision wastage of the existing methods by considering all potential positive samples to optimize the model which outperforms other metric learning losses on LSVGL.

The paper is organized as bellow: related works with more details are presented in section \ref{sec:related_work}. Then we give a briefly overview of the proposed method and a detailed introduction in section \ref{sec:overview} and \ref{sec:method}, respectively. Datasets and experiments are presented in section \ref{sec:experiments}. Conclusions and outlook are presented in section \ref{sec:conclusion}.

\section{Related Work}\label{sec:related_work}
In this section, we briefly review the methods related to LSVGL. First, we summarize the development of CNNs, and then list the research of metric learning and LSVGL.

\textbf{Deep CNNs.}
Many popular deep networks including AlexNet, VGGNet, GoogleNet, and ResNet networks have been proposed in recent years. Many works have proved that the VGG network pre-trained on the ImageNet image classification dataset is better than other baseline models in the LSVGL task. Therefore, in this paper, we also choose VGG16 as the benchmark network like most other papers. VGGNet is a deep convolutional neural network developed by Oxford University's Visual Geometry Group and researchers from Google DeepMind. Its main contribution is to show that the depth of the network is a key part of the excellent performance of the algorithm. Based on the VGGNet model, many excellent models have been improved, such as the use of new optimization algorithms and multi-model fusion. Furthermore, recognition of salient objects in a scene is helpful to extract the fixed and unchanging structural features at a certain location, the image representation extracted by VGGNet are used in many image recognition tasks\cite{multi-salient, hybrid-graph,contour-transfer18, webly-salient-20}. Yang \cite{multi-salient} proposed a multi-scale cascade network to identify the most visually conspicuous objects in an image and Luo \cite{hybrid-graph} presented a hybrid graph neural network to interweaving the multi-scale features for crowd density as well as its auxiliary task together, to better capture and consolidate multi-scale high-level context information for object skeleton detection, \cite{contour-transfer18} automatically converted an existing deep contour detection model into a salient object detection model without using any manual salient object masks by grafting a new branch onto a well-trained contour detection network. Luo \cite{webly-salient-20} proposed a webly-supervised learning method for salient object detection with no pixel-wise annotations.

\textbf{Deep metric learning.}
Metric learning is a distance function that measures similarity: similar objects are close to each other, and dissimilar ones are far away. While deep learning is still asleep, the Mahalanobis distance in Euclidean space solves the judgment of distance in traditional metric learning, Cross-view Quadratic Discriminant Analysis (XQDA) and Keep It Simple and Straightforward Metric learning (KISSME) were both classic metric learning methods in the part. Deep metric learning currently mainly uses the network to extract embedding, and then uses L2-distance to measure distances in the embedding space.

In recent years, a large amount of work has applied metric learning from person re-identification to LSVGL \cite{arandjelovic2016netvlad, liuliu19, kim2017learned, zhu2018attention}. Two images at the same location are defined as a positive pair, while two images at different locations are defined as a negative pair. Triplet loss has been used with great success in LSVGL \cite{arandjelovic2016netvlad, zhu2018attention}. As an upgraded version of triplet loss, the triplet hard sample mining loss (TriHard) selects the hardest negative sample, even if the anchor is farthest from the nearest (hardest) negative sample. Quadruplet loss extends TriHard loss, retains the idea of 'hardest', and simultaneously selects one more negative sample and adds the absolute distance between the negative samples to the training. Margin Sample Mining Loss (MSML) is a combination of the advantages of the above metric learning methods. It selects the most similar positive samples (the hardest) and the most similar negative samples (the hardest) to optimize the quadruplet loss. Unfortunately, TriHard, quadruplet, MSML did not verify performance in LSVGL, in this paper we make up for this regret.

\textbf{Deep LSVGL related methods.}
LSVGL remains a challenging task in urban environments due to the repetitive structures, illumination condition, and differences in perspective. While there have been many improvements in designing better LSVGL systems \cite{arandjelovic2016netvlad, zhu2018attention, kim2017learned, liuliu19, chen2017deep}. Arandjelovic \cite{arandjelovic2016netvlad} performed learning for LSVGL and proposed the NetVLAD representations, which significantly outperform the previous local-feature-based representations on several benchmarks. Zhu \cite{zhu2018attention} proposed an attention mechanism to solve the LSVGL task. However, none of these works actually training the CNNs for the task considered anchors are almost assigned into multiple images, and these images should be used as positive samples to network learning.

\begin{figure*}
	\centering
	\includegraphics[scale=0.98]{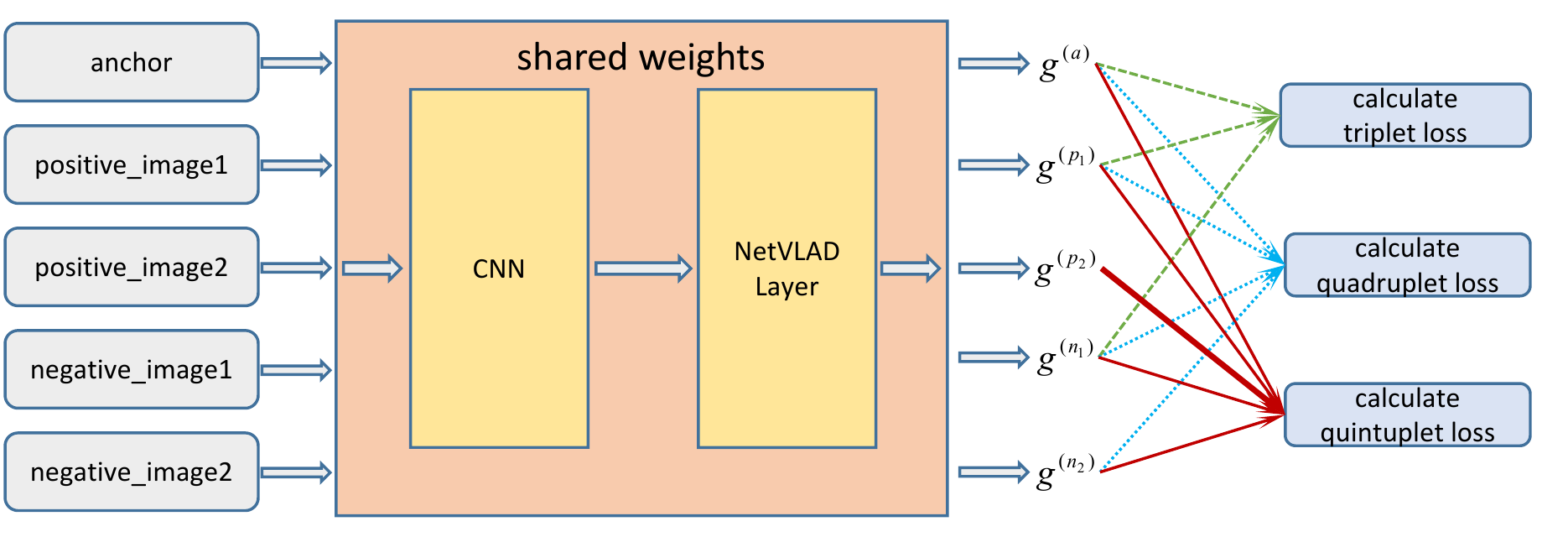}
	\caption{A schematic diagram of metric learning in deep neural networks.}
	\label{fig:metric_learning_diagram}
\end{figure*}

\section{An Overview of Our Method}\label{sec:overview}
We briefly introduce the overview of the proposed method. We maintain the view that the current methods to solve the LSVGL tasks have two main shortcomings: (1). The metric learning methods used for LSVGL is too limited. Almost the current state-of-the-art methods use triplet loss, however many excellent metric learning methods have already been proposed, such as quadruplet loss, TriHard, MSML methods, etc. and (2). We can summarize the LSVGL problem into the task of image retrieval in urban environments. Although it is the same location, it is often taken during retrieval compared with the image used in training the geometric database, the image of this place will appear infection factors such as lighting, perspective, occlusion, and the dynamic scene that everyone knows. Many works are also working on this kind of problem and have achieved quite good results, but there is almost no work to focus on during the training phase, anchor exists in multiple positive samples, and the most commonly used triplet loss only picks the nearest positive sample to learn, which undoubtedly reduces the constraints of positive examples on anchor. Fig. \ref{fig:metric_learning_diagram} shows a schematic diagram of metric learning in deep neural networks.

To address the first shortcoming, we first summarize the current mainstream metric learning methods, and use the current state-of-the-art LSVGL network NetVLAD to compare the impact of multiple metric learning methods on geo-localization performance.

To address the second shortcoming, we considered the characteristics of LVSGL, that is, the problem caused by the arbitrariness of the perspective that anchors may exist in multiple positive samples, we propose a new metric learning method called QUITLoss. QUITLoss is originally designed to expand only one positive sample into multiple positive samples based on triplet loss. We use QUITLoss to correct each metric learning method and compare the performance of other metric learning methods after correction.

\section{Proposed Quintuplet Loss}\label{sec:method}
First, we introduce several metric learning methods, including triplet loss, TriHard loss, quadruplet loss, and MSML loss. And then we formally introduce the proposed QUITLoss.

\subsection{Background}
To extract image depth embedding, metric learning aims to learn a mapping: $g(x): R^{W \times H} \rightarrow R^D$ which maps semantically similar images from the raw data manifold in $R^{W \times H}$ onto metrically close points in $R^D$ \cite{trihard17}, where $W$ and $H$ are the width and height of input image respectively and $D$ is the dimension of the final embedding. The deep metric learning aims to find the mapping through minimizing the metric loss of training data. Then a metric function $L(x,y): R^D \times R^D \rightarrow R$ is designed to measure similarity in the embedding space.

\textbf{Triplet Loss}. Triplet loss generates more distinctive features than softmax loss \cite{facenetSchroffKP15}. A triplet in triplet loss contains three different images $\{I^{(a)}, I^{(p)}, I^{(n)}\}$, while $I^{(a)}$ is the anchor image and $I^{(p)}$ and $I^{(n)}$ represent the positive image in the same place with $I^{(a)}$ and negative image in another different place with $I^{(a)}$ respectively. A $W$-by-$H$ image generates a $D$-dimensional feature after network mapping, and a triplet of features $\{g^{(a)}, g^{(p)}$, $g^{(n)}\}$ would be embedded into the triplet loss formulation as bellow:
\begin{equation}
L_{tri}=h(\overbrace{||g^{(a)}-g^{(p)}||_2}^{to\ shorten} - \overbrace{||g^{(a)}-g^{(n)}||_2}^{to\ enlarge} + \alpha)
\label{eq:triplet}
\end{equation}

where $h(x)=max(x,0)$ and $\alpha$ is the margin to distinguish between the positive samples and the negative ones. The first term $||g^{(a)}-g^{(p)}||_2$ shortens the distance between $I^{(a)}$ and $I^{(p)}$, and the second term $||g^{(a)}-g^{(n)}||_2$ enlarges the distance between $I^{(a)}$ and $I^{(n)}$.

\textbf{Quadruplet loss.} Compared with the triplet loss, quadruplet loss \cite{quadruplet17} contains one more negative. A quadruplet selects four different images $\{I^{(a)}, I^{(p)}, I^{(n_1)}, I^{(n_2)}\}$, where $I^{(a)}$ and $I^{(p)}$ have the same meaning as triplet loss, $I^{(n_1)}$ and $I^{(n_2)}$ are negative samples. The quadruplet loss is formulates as bellow:
\begin{equation}
\begin{aligned}
L_{quad}&=h(\overbrace{||g^{(a)}-g^{(p)}||_2 - ||g^{(a)}-g^{(n_1)}||_2 + \alpha}^{relative\ distance}) \\
&+ h(\overbrace{||g^{(a)}-g^{(p)}||_2 - ||g^{(n_1)}-g^{(n_2)}||_2 + \beta}^{absolute\ distance})
\end{aligned}
\label{eq:quadruplet}
\end{equation}
where $\alpha$ and $\beta$ are the margin in two terms.  The first term is a triplet loss, which focuses on the relative distance between positive pair and negative pair, the second term adds the absolute distance by considering different negative images.

\textbf{TriHard Loss}. As the possible number of quadruplets grows rapidly as the data volume gets larger, loss Eq. (\ref{eq:quadruplet}) can be further optimized. In order to relieve this,  triplet loss with hard sample mining computes a batch of samples together. For each image, the TriHard loss picks the most similar sample with different place $I^{(h,n)}=argmin_n (||g^{(a)}-g^{(n)}||_2)$ to generate a triplet $\{I^{(a)}, I^{(p)}, I^{(h,n)}\}$, then estimate loss by Eq. (\ref{eq:triplet}).

\textbf{Margin Sample Mining Loss (MSML).} MSML picks the most dissimilar positive pairs and most similar negative pair in the whole batch, as:
\begin{equation}
L_{msml}=h(||g^{(a)}-g^{(h,p)}||_2-||g^{(h,n_{1})}-g^{(h,n_{2})}||_2+\alpha)
\end{equation}

where $g^{(h,n_{1})}$ and $g^{(h,n_{1})}$ can share the same identity with $I^{(a)}$ or not. 

\subsection{QUITLoss}

\subsubsection{Definition}
\begin{figure}[h]
	\centering
	\includegraphics[scale=0.45]{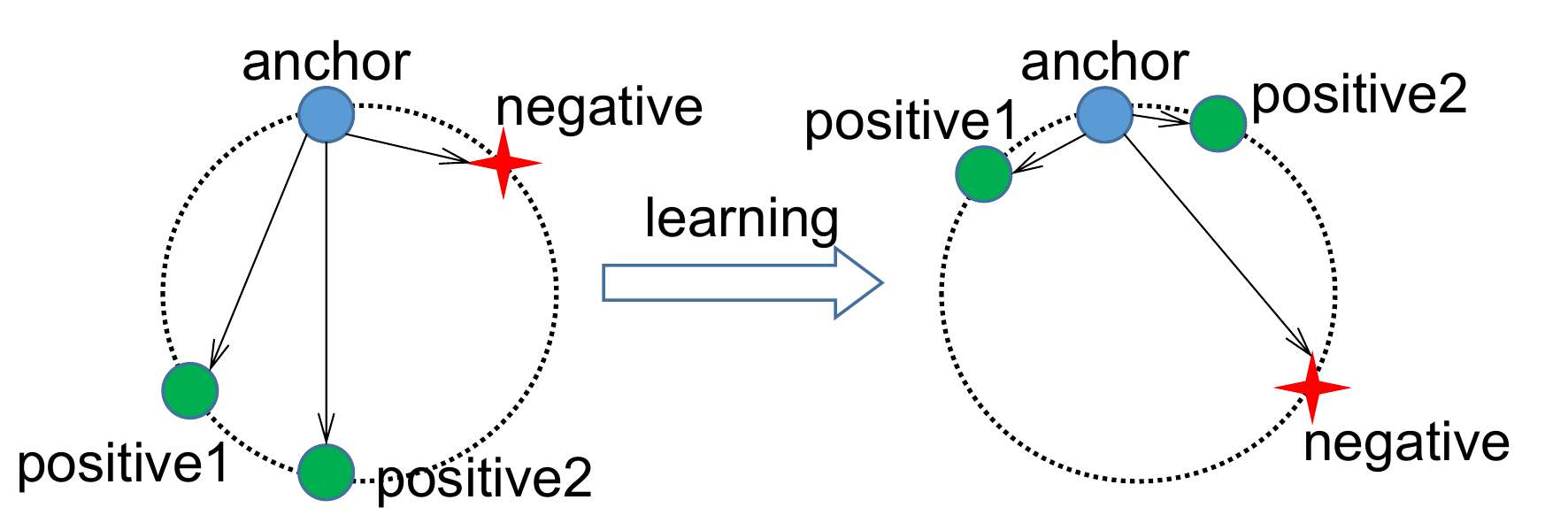}
	\caption{QUITLoss keeps the same learning strategy with triplet loss but selects multiple positives.}
	\label{fig:quit-loss-graph}
\end{figure}

We apply a new positive example mining strategy for LSVGL named QUITLoss. Different from the loss introduced above, QUITLoss selects multiple positive samples to integrate into the final loss calculation:
\begin{equation}
L_{quit}=\sum_{i=1}^k h(||g^{(a)}-g^{(p_{(i)})}||_2 - \mathcal{P} + \alpha)
\end{equation}
where $g^{(p_{(i)})}$ is the feature of the $i$-th positive of the anchor image $I^{(a)}$, and $\mathcal{P}$ represents multiple negative sample selection methods, which will be explained in detail below. Fig. \ref{fig:quit-loss-graph} shows the learning process.

We design QUITLoss as a general concept, not a specific mathematical form. Different negative sample selection strategies can form specific expressions. 
When $\mathcal{P}=||g^{(a)}-g^{(h, n)}||_2$, the final QUITLoss can be formed after fusion with the TriHard loss:
\begin{equation}
L_{quit\_trihard}=\sum_{i=1}^k h(||g^{(a)}-g^{(p_{(i)})}||_2 - ||g^{(a)}-g^{(h, n)}||_2 + \alpha)
\label{eq:trihard-quit}
\end{equation}
where $g^{(h, n)}$ is the feature of the nearest negative sample of the anchor image $I^{(a)}$.

We set $\mathcal{P}=||g^{(a)}-g^{(n_1)}||_2$ and rewrite the mathematical expression after the fusion of QUITLoss and quadruplet loss:
\begin{equation}
\begin{aligned}
L_{quit\_quad}&=\sum_{i=1}^k h(||g^{(a)}-g^{(p_{(i)})}||_2 - ||g^{(a)}-g^{(n_1)}||_2 + \alpha) \\&+ \sum_{i=1}^k h(||g^{(a)}-g^{(p_{(i)})}||_2 - ||g^{(n_1)}-g^{(n_2)}||_2 + \beta)
\end{aligned}
\end{equation}

\subsubsection{Gradient}
In addition, we also provide the gradient information of TriHard loss modified by QUITloss. The gradient of the loss function $L_{quit\_trihard}$ for each variable is calculated as follows:
\begin{equation}
\begin{aligned}
\frac{\partial L}{\partial {g^{(a)}}} &= \sum_{i=1}^k 2(g^{(a)}-g^{(p_{(i)})} - (g^{(a)}-g^{(h,n)})) \\&=\sum_{i=1}^k 2(g^{(h,n)} - g^{(p_{(i)})})
\end{aligned}
\end{equation}

\begin{equation}
\begin{aligned}
\frac{\partial L}{\partial g^{(p_{(i)})}} &= 2(g^{(a)}-g^{(p_{(i)})})(-1) =2(g^{(p_{(i)})}-g^{(a)}), \\& i=1, 2, ..., k
\end{aligned}
\end{equation}

\begin{equation}
\frac{\partial L}{\partial g^{(h, n)}} = -2(g^{(a)}-g^{(h, n)})(-1)=2(g^{(a)}-g^{(h, n)})
\end{equation}
Other loss forms can be deduced in the same way.

In summary, compares with other metric learning losses, our QUITLoss has the following advantages. First, QUITLoss makes up for the problem that the metric learning methods in the current LSVGL task do not consider the anchor images assigned to multiple positive samples caused by the difference in perspective. Second, QUITLoss does not completely discard the existing methods, but provide a general idea, which can be extended to the existing metric learning loss. Finally, QUITLoss is easy to implement.

\begin{table*}[t]
\renewcommand\arraystretch{1.5}
  \centering
  \caption{Comparison of state-of-the-art deep metric learning loss. All the results are from the 512-D representations based on VGG-16 architecture.}
  \label{tab:performance_comparison}
  \setlength{\tabcolsep}{4mm}{
    \begin{tabular}{|c|c|c|c|c|c|c|c|}
    \hline
    \multirow{2}{*}{Base model} &
    \multirow{2}{*}{Methods} &
    \multicolumn{3}{c|}{Pitts250k-test} & \multicolumn{3}{c|}{Tokyo 24/7}\cr\cline{3-8}
    & &Recall@1&Recall@5&Recall@10&Recall@1&Recall@5&Recall@10\cr
	\hline
     \multirow{8}{*}{VGG16}& NetVLAD\cite{arandjelovic2016netvlad} &0.8066&0.9088&0.9306 & 0.6000 & 0.7365& 0.7905\cr\cline{2-8}
    & triplet& 0.8169 &  0.9112 &  0.9329 &0.6286 & \textbf{0.7714}& 0.8032\cr
    & triplet + QUITLoss (ours)& \textbf{0.8226} & \textbf{0.9118} & \textbf{0.9341} & \textbf{0.6349}  & 0.7587 & \textbf{0.8159}\cr\cline{2-8}
    & quad& 0.7950 &  0.8966 &  0.9209 & \textbf{0.6508}& 0.7841& 0.8317\cr
    & quad+QUITLoss (ours)& \textbf{0.8030} & \textbf{0.9010} &  \textbf{0.9252} & 0.6476 & \textbf{0.7905} & \textbf{0.8413} \cr\cline{2-8}
    & trihard& 0.8333 &0.9203&0.9313& 0.6730& 0.7905 & 0.8381 \cr
    & trihard + QUITLoss (ours)& \textbf{0.8403} & \textbf{0.9216} & \textbf{0.9401} & \textbf{0.6952} & \textbf{0.8159} & \textbf{0.8476} \cr\cline{2-8}
    & MSML& 0.7431 & 0.8639 & 0.8958 & 0.6032& 0.7492& 0.8032\cr\cline{2-8}
    \hline
    \end{tabular}}
\end{table*}

\section{Experiments}\label{sec:experiments}

\subsection{Datasets and Evaluation Methodology}
We report results on two publicly available datasets. \textbf{Pittsburgh (Pitts250k)} \cite{torii2013visual}  is a popular LSVGL benchmark and contains 250k database images downloaded from Google Street View and 24k queries generated from Street View but taken at different times, years apart.  Similar to \cite{arandjelovic2016netvlad}, we divide Pitts250k into three roughly equal parts for training, validation, and testing, namely Pitts250k-train, Pitts250k-valid, and Pitts250k-test respectively. Each part contains approximately 8k training images and 8k queries. To facilitate faster training, we use a smaller subset (Pitts30k) which contains 10k training images for training.

\textbf{Tokyo 24/7} \cite{torii201524} is another extremely challenging dataset which contains 76k database images and 315 queries.  The database images were only taken at daytime from Google Street View as described above, but the queries were taken under extreme illumination conditions.

\textbf{Evaluation metric.} We follow the standard LSVGL evaluation procedure \cite{retrieval12, torii201524, torii2013visual, arandjelovic2016netvlad}. The localization is deemed correct if at least one of the top N retrieved database images are within $d$=25 meters from the query. $Recall@N$ is the percentage of correctly localized queries for different values of N.

\begin{figure*}[t]
	\centering
	\includegraphics[scale=0.56]{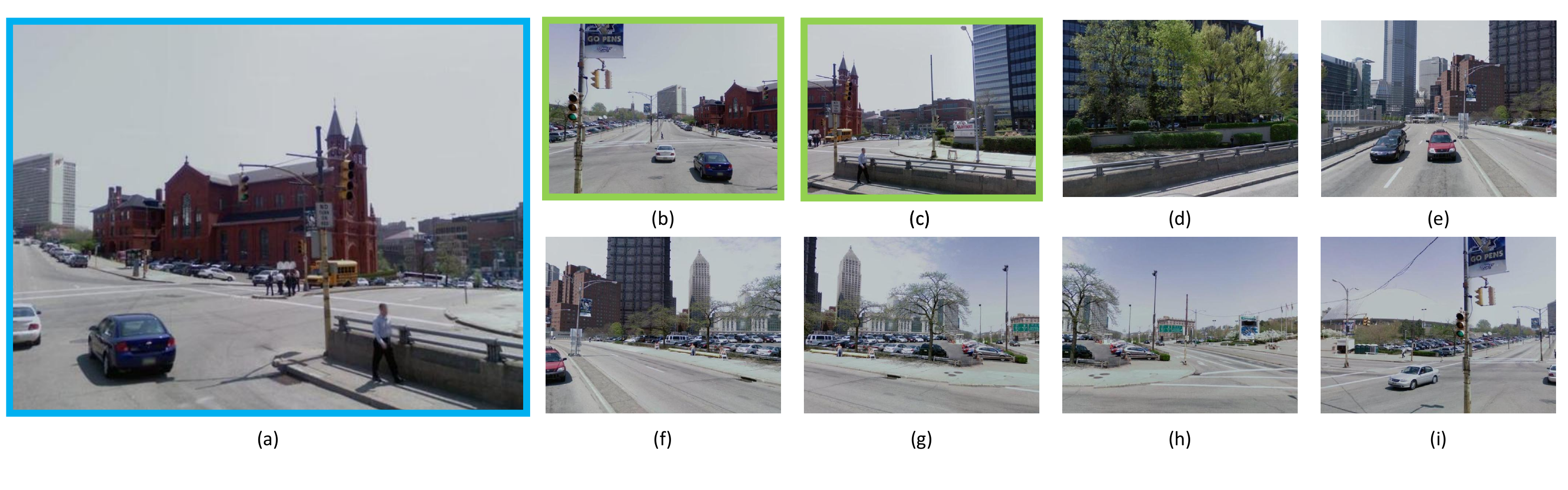}
	\caption{Schematic diagram of positive sample selection in QUITLoss. (1). Image (a) is the anchor image, and image (b)-(i) share the same geometric location with the anchor (a). In our experiment, image (b)-(i) are considered as potential positives, and other metric learning losses only choose the nearest image (b) as the positive,  but the brick red church exists in multiple images (b) and (c), we choose two positives to our QUITLoss. (2). If our QUITLoss keeps the hardest positive selection strategy, It is entirely possible to pick out (d)-(i), those are certainly not positive samples from the perspective of human supervision.}
	\label{fig:not_hardest_positive}
\end{figure*}

\subsection{Implementation Details}
Each image is resized into 224 $\times$ 224 pixels with data augmentation. The data augmentation includes randomly flipping, zooming, blurring, and contrast enhancement. We use the VGG16 pre-trained model as our base model, the margin $\alpha$ and $\beta$ are set to 0.3 and 0.2 respectively. During gradient descent, the SGD optimizer is used and the initial learning rate is set to $10^{-4}$ and drops 0.5 times every 5 epochs. The training is terminated if the performance is not improved for 10 consecutive epochs.

\begin{table}[t]
\renewcommand\arraystretch{1.2}
  \caption{Experimental results on the influence of different hyperparameter $k$ on the results of geographical location.}
  \label{tab:learning_k}
  \setlength{\tabcolsep}{6mm}{
    \begin{tabular}{|c|c|c|c|}
    \hline
    \multirow{2}{*}{$k$}&
    \multicolumn{3}{c|}{Pitts250k-test}\cr\cline{2-4}
    &Recall@1&Recall@5&Recall@10\cr
    \hline
    1 & 80.66 &  90.88 & 93.06\cr\hline
    2 & \bf{82.34} & \bf{90.94} & \bf{93.41} \cr\hline
    3 & 81.64 & 90.75 & 93.18 \cr\hline
    4 & 74.79 & 86.28 & 89.93 \cr\hline
    \end{tabular}}
\end{table}

\textbf{Learning the hyperparameter k.} To begin with our experiments, we first study the impacts of the hyperparameter k of our method. We set different parameters to verify the impact of different numbers of positive samples on the Pittks250k-test dataset. The network parameter is the baseline network when $k$=1. The experiment found that the network does not converge when $k$ is greater than 4, and the localization results are best when $k$ = 2, which also meets the theoretical requirements. Table \ref{tab:learning_k} shows the details. The same observation can be seen on the Tokyo 24/7 dataset. The experimental results also verify our cognition of LSVGL, that is, due to the arbitrariness of the two shooting angles, the anchor image is almost allocated to two positive samples. Thus, we empirically select k=2 in all our following experiments.

\subsection{Results and Discussion of Different Losses}
We conduct experiments with different losses and provide the results to illustrate the effectiveness of the proposed QUITLoss. The PyTorch version of the open-source code \cite{arandjelovic2016netvlad} is used in our experiment and four metric learning losses, triplet loss, quadruplet loss, TriHard loss, MSML are implemented to compare with the origin triplet loss.


\textbf{Results of Different losses.} 
In our experiments, we compared the impact of different metric learning methods on geo-localization performance. The experimental results show that TriHard has advantages over other metric learning methods because it considers using the  nearest (hardest) negative samples to constrain anchors in the metric space. On the Pitts250k-test dataset and under the recall @ 1 indicator, the TriHard method is 2.67\% higher than the state-of-the-art method NetVLAD and is 1.64\% higher than the triplet method we implemented. Quadruplet loss does not significantly help the localization results, because although quadruplet loss uses the absolute distance between two negative samples to constrain the anchor, it still does not have a strong constraint on the anchor with the nearest negative sample. It is worth mentioning that the training of MSML has not been successful, and the performance of MSML on the localization results is greatly reduced. The reason why the hardest (farthest) positive sample cannot be selected will be analyzed later. On the Tokto247 dataset, we also have the same conclusion, see Table \ref{tab:performance_comparison} for details.

\textbf{Discussion of QUITLoss.}
Every metric learning method we use can be corrected with QUITLoss. Experimental results show that the corrected metric learning methods can improve the performance of geo-localization. Similarly, the corrected TriHard loss has better localization results than other metric learning methods. After correction, the triplet loss, quadruplet loss, and TriHard are improved by 0.57\%, 0.80\%, and 0.70\%, respectively, compared to the performance of the methods before correction. The performance reported on Tokyo247 is shown in Table \ref{tab:performance_comparison}.

\textbf{Comparison with state-of-the-art.}
NetVLAD uses Pytorch's built-in functions to calculate triplet loss. For an anchor, multiple negative samples are considered. Taking TriHard as an example, the performance of pure TriHard loss is 2.67\% higher than NetVLAD, and TriHard loss is 3.37\% higher after QUITLoss correction.

\textbf{Why can we not choose the hardest positive?} In LSVGL system, we can't choose the most difficult positive samples, because in panoramic images, we take those images which are close to the anchor image's geographical location as potential positive samples, but there are only two positive samples (we have proved why there are only two) that really belong to the anchor, so in LSVGL only the nearest (not the most difficult) can be selected. As Table \ref{tab:performance_comparison} shows, MSML loss significantly degrades performance, and Fig. \ref{fig:not_hardest_positive} shows that the most difficult is not the anchor's positive sample, and there is no semantic similarity between the most difficult and the anchor. For example, the images that are taken from the opposite perspective proves this conclusion.

\section{Conclusion}\label{sec:conclusion}
In this paper, we propose a new metric learning loss with multiple positive constraints named quintuplet loss for Large Scale Visual Geo-Localization (LSVGL). In our method, we calculate a distance matrix and choose the smallest two distance of positive pairs to form a general strategy. In this way, quintuplet loss used all the positives to train the model. In addition, we also carefully analyze why selecting two positive samples is the best strategy.

We use the pre-trained VGG-16 model as our baseline and the state-of-the-art model NetVLAD to do contrast experiments with different metric learning losses on several benchmarks datasets, including Pitts250K, Tokyo24/7. The results show the metric learning loss corrected by our quintuplet loss gets better performance and learns a finer metric in feature embedding space.

\ifCLASSOPTIONcompsoc
  \section*{Acknowledgments}
\else
\fi
The authors gratefully acknowledge the financial supports by the National Science Foundation of China under grant numbers U1964203, useful suggestions given by Xin Li, Fan Yang, Ao Luo, Hao Guo are also acknowledged.

\ifCLASSOPTIONcaptionsoff
  \newpage
\fi



\bibliographystyle{unsrt}
\bibliography{reference}   

\begin{IEEEbiography}
[{\includegraphics[width=1in,height=1.25in,clip]{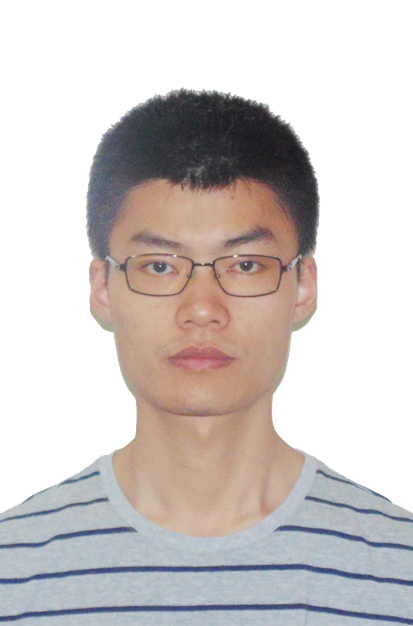}}]{Qiang Zhai} received Master degree from Beijing Jiaotong University in 2014. Now he is the Ph.D candidate in Pattern Recognition and Machine Intelligence Lab, UESTC. His current research interests include visual place recognition, SLAM, deep learning.
\end{IEEEbiography}

\begin{IEEEbiography}
[{\includegraphics[width=1in,height=1.25in,clip]{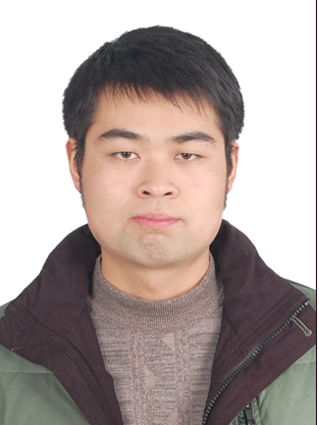}}]{Rui Huang} is a post-doctor in School of Automation Engineering, Center for Robotics, University of Electronic Science and Technology of China (UESTC). He received Ph.D degree in Control Science and Engineering from UESTC in July 2018. Dr. Huang has been a joint training doctoral student in TAMS, University of Hamburg from 2016 to 2017. His current research interests include reinforcement learning, exoskeleton, human-robot interaction, robot control. 
\end{IEEEbiography}

\begin{IEEEbiography}
[{\includegraphics[width=1in,height=1.25in,clip]{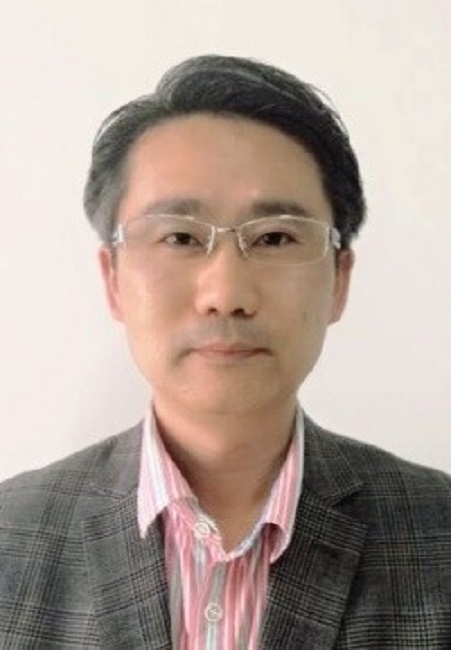}}]{Hong Cheng} is a full Professor in School of Automation, a vice director of Center for Robotics, UESTC. He received Ph.D degree in Pattern Recognition and Intelligent Systems from Xi'an Jiaotong University in 2003. Now he is the founding director of Machine Intelligence Institute, UESTC. Before this, he was a postdoctoral at Computer Science School, Carnegie Mellon University, USA from 2006 to 2009. He was an associate Professor of Xi'an Jiaotong University since 2005. Since July 2000, he had been with Xi'an Jiaotong University, where he had been a team leader of Intelligent Vehicle Group at the Institute of Artificial Intelligence and Robotics before going to USA. Dr. Cheng has been a senior member of IEEE, ACM.
\end{IEEEbiography}


\begin{IEEEbiography}
[{\includegraphics[width=1in,height=1.25in,clip]{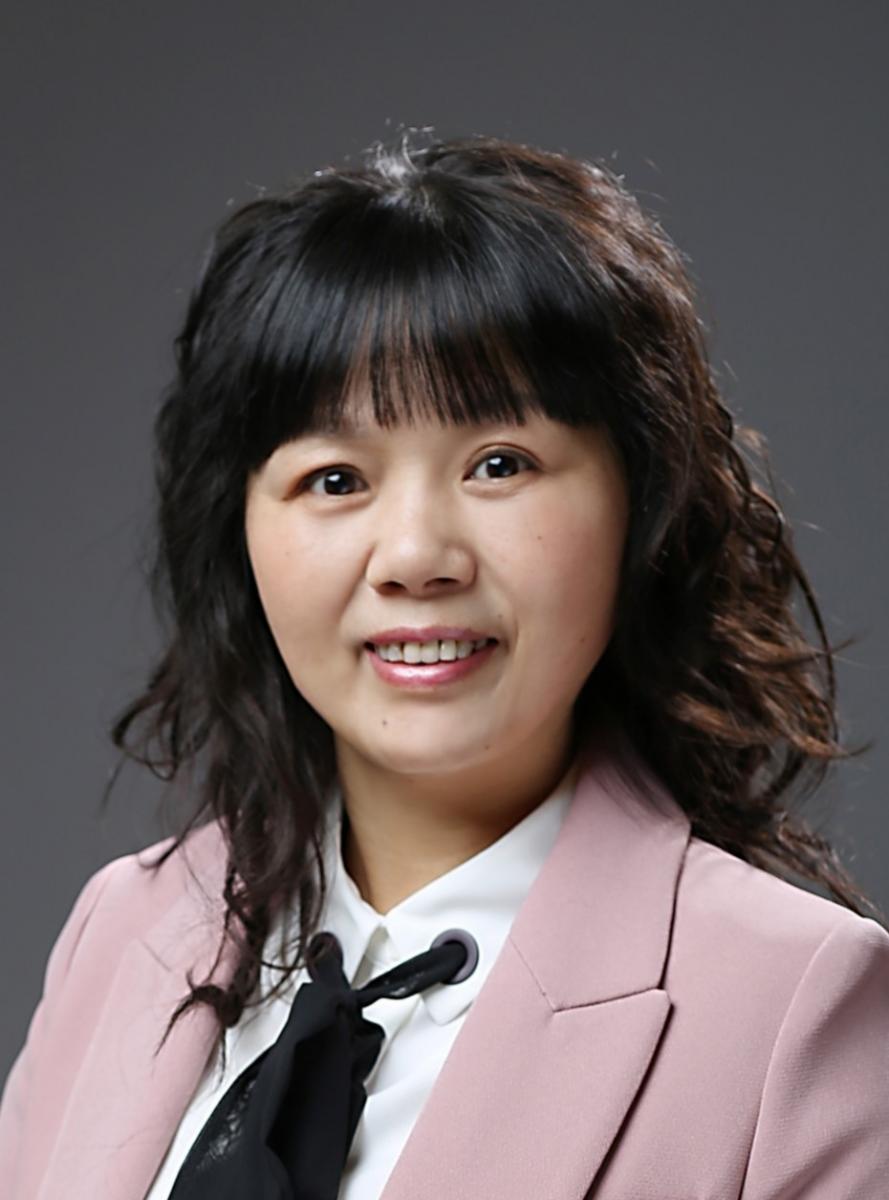}}]{Huiqin Zhan} is a full Professor in School of Automation, UESTC. After graduated from BUAA , she has worked in Chengdu Aeronautic Instrument Company for 9 years, engaged in new products design. During that period, she has been involved in several projects, including computer of airplane atmospheric data, high accuracy digital stress sensor, and sensor automatic test system. She has worked in a university since 1993 and her research interest focused on modern test theory and methodology, virtual instrument, automatic measurement, computer  automatic control technology,  signal processing, and system modeling.
\end{IEEEbiography}

\begin{IEEEbiography}
[{\includegraphics[width=1in,height=1.25in,clip]{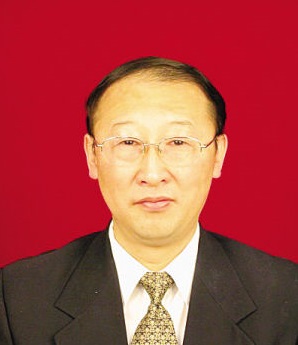}}]{Jun Li} is an academician of Chinese Academy of Engineering and a full  professor of school of vehicle and transportation, Tsinghua University. He received the PhD degree from Jilin University of Technology, in 1989 and has broken through four core technologies of automobile engine, namely, design, combustion, electric control and reliable durability, and built a basic technology research platform. Presided over the independent research and development of heavy-duty series diesel engines to make FAW Jiefang Truck heavy-duty. He presided over and completed 10 863 and 973 subjects, published 95 papers and 1 monograph, and won the first and second prizes of national science and technology progress, and 6 provincial and ministerial first prizes.
\end{IEEEbiography}

\begin{IEEEbiography}
[{\includegraphics[width=1in,height=1.25in,clip]{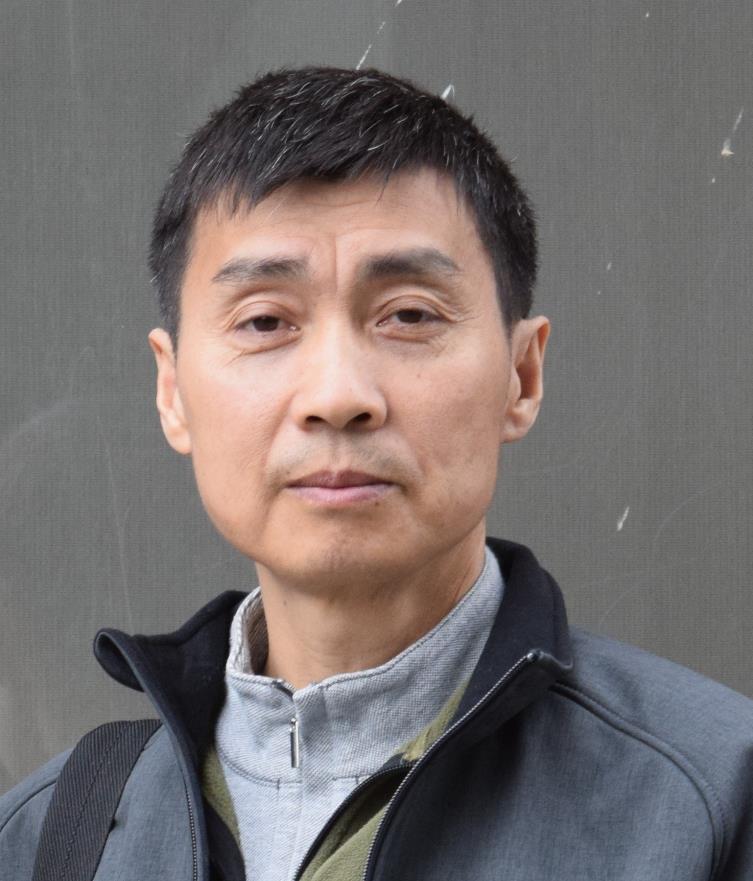}}]{Zicheng Liu} received the BS degree from Huazhong Normal University, China, in 1984, the MS degree from Institute of Applied Mathematics, Chinese Academy of Sciences, in 1989, and the PhD degree in computer science from Princeton University, in 1996, He is a principal researcher with Microsoft Research Redmond. Before joining Microsoft Research, he worked with Silicon Graphics Inc. His current research interests include human activity recognition, 3D face modelling, and multimedia signal processing. Dr. Liu has been a fellow of the IEEE.
\end{IEEEbiography}

\end{document}